\documentclass[conference]{IEEEtran}
\IEEEoverridecommandlockouts

\usepackage{cite}
\usepackage{amsmath,amssymb}
\usepackage{graphicx}
\usepackage{xcolor}
\usepackage{tikz}
\usetikzlibrary{arrows.meta, positioning}
\usepackage{booktabs}
\usepackage{makecell}
\usepackage{algorithm}
\usepackage{algorithmic}
\renewcommand{\thealgorithm}{\arabic{algorithm}}

\def\BibTeX{{\rm B\kern-.05em{\sc i\kern-.025em b}\kern-.08em
T\kern-.1667em\lower.7ex\hbox{E}\kern-.125emX}}

\begin{document}
\title{
Adaptive Two-Stage Online Learning for Service-Affecting Failure Detection in\\Mobile Core Networks
}
\author{
\IEEEauthorblockN{J.~du~Toit$^{*\dagger}$, G.~Fita$^{*}$, J.~Salzwedel$^{\ddagger}$, A.~Stoltz$^{\dagger\S}$, R.~Wolhuter$^{*}$}

\IEEEauthorblockA{
\textit{$^{*}$Department of Electrical and Electronic Engineering, Stellenbosch University, South Africa}\\
\textit{$^{\S}$School of Computer Science and Applied Mathematics, University of the Witwatersrand, South Africa}\\
\textit{$^{\dagger}$Technology Strategy Planning, Architecture \& Assurance, Vodacom Group Limited, South Africa}\\
\textit{$^{\ddagger}$Core Data Networks Performance Management, Vodacom Group Limited, South Africa}\\
\{jacowp357, gospelantoniofita, jasonsalzwedel\}@gmail.com, Adham.Stoltz@vodacom.co.za, wolhuter@sun.ac.za
}
}

\maketitle

\begin{abstract}
Mobile network operators monitor aggregated traffic volumes to assess the operational
health of core network infrastructure. Reliable failure detection is challenging due to strong temporal structure,
non-stationarity, measurement artefacts, and extreme class imbalance, which limit
static threshold-based monitoring.

This paper proposes a two-stage online learning framework for traffic-based failure
detection in mobile core networks. Stage~I incrementally models normal traffic dynamics
using lightweight regression with time-aware features. Stage~II analyses prediction
residuals together with contextual indicators to detect genuine service-affecting network failures.
The framework operates fully online under a prequential evaluation protocol, enabling
continuous adaptation with low computational overhead.

Across linear and non-linear models, the proposed two-stage architecture achieves
the best precision--recall trade-off, attaining the highest recall, F1-score, and
AUC at acceptable false positive rates. These results demonstrate the importance of
explicit residual decomposition for reliable failure detection in streaming mobile core network data.
\end{abstract}


\section{Introduction}
\label{sec:introduction}

Mobile network operators (MNOs) rely on continuous monitoring of aggregated traffic
volumes to assess the health of their core networks. The mobile core network (MCN)
comprises the packet-switched control and user-plane infrastructure responsible for
session management, mobility, authentication, policy enforcement, and routing of user
traffic between the radio access network (RAN) and external data networks. Traffic from multiple radio access technologies, access point names (APNs), and
core network elements is aggregated per geographical region at fixed intervals,
providing a scalable and technology-agnostic view of system behaviour.

Although aggregated traffic exhibits strong diurnal and weekly structure driven by human
activity~\cite{xu2016understanding}, it is inherently non-stationary and subject to
measurement artefacts such as counter overflows, partial capture,and delayed reporting. These effects are not necessarily pervasive across all MNOs or deployments. However, core network equipment commonly relies on incremental traffic counters with finite bit-width, which can saturate or fail under sustained high load. Service continuity is often maintained through redundancy or backup processing
paths that remain transparent to subscribers, while counter information is lost or corrupted. This results in abrupt apparent traffic drops that closely resemble genuine network failures, undermining static threshold-based monitoring leading to high false alarm rates.

Traditional threshold-based systems remain widely used due to their simplicity, but they
assume quasi-stationary behaviour and require continuous manual tuning. Predictable
contextual effects such as weekends, holidays, and tariff changes frequently trigger
false alarms, eroding trust in automated decision-making systems~\cite{Lakhina2004Characterization,Chandola2009Anomaly}.
As networks scale in complexity and load, these limitations increasingly constrain
their operational usefulness.

Prior work has explored statistical and learning-based approaches to traffic anomaly
detection, including residual analysis, change-point detection, and deep learning
models~\cite{Basseville1993Detection,Laptev2015Generic,Malhotra2015LSTM,Ren2019TimeSeries}.
However, most approaches are trained offline and treat all deviations as a single
anomaly class, conflating service-impacting failures with benign measurement artefacts.
This limits robustness, interpretability, and operational trust in MCN environments. 

Network monitoring data is inherently generated as a data stream, motivating online
learning approaches that adapt incrementally to evolving traffic patterns~\cite{ShalevShwartz2012Online,Gama2014Survey}. While recent work has explored online
anomaly detection for networked systems~\cite{burgueno2020online,shahraki2022comparative},
most methods remain end-to-end and do not explicitly separate baseline traffic dynamics
from failure detection.

In contrast, this work proposes a two-stage online learning architecture that decouples
self-supervised traffic baseline modelling from rare-event failure detection. By
isolating predictable traffic dynamics before classification, the framework improves
robustness under severe class imbalance, delayed labels, and measurement artefacts,
while remaining computationally lightweight and operationally aligned.

The proposed approach is intended to complement, rather than replace, existing core network failure detection systems. Some failure modes may not manifest clearly in aggregated traffic and are therefore better captured by more granular monitoring tools. Conversely, issues originating outside the mobile core network, such as failures in external service platforms, may still be reflected in aggregate traffic and can be detected by the proposed method. Importantly, aggregated traffic is a primary revenue-generating KPI for MNOs, and any service-affecting disruption visible in this signal has direct commercial impact. Even modest reductions in detection latency can therefore translate into meaningful operational and revenue benefits.

The main contributions of this paper are:
\begin{itemize}
\item A two-stage online architecture separating baseline modelling and failure detection.
\item Evaluation of time-aware feature encodings under streaming constraints.
\item Demonstration of strong performance and stability using lightweight online models
on real operational MCN data.
\end{itemize}

This work was carried out in collaboration with Vodacom Group Limited as part of an ongoing continuous improvement programme within the MCN performance management domain.

\section{Problem Formulation}
\label{sec:problem}

\subsection{Operational Interpretation of Traffic Anomalies}
\label{subsec:definitions}
In aggregated MCN monitoring, deviations from expected traffic
volumes may arise from fundamentally different causes. In this work, we
distinguish between two primary sources of anomalous behaviour: genuine network
failures and measurement artefacts.

A \emph{network failure} corresponds to any condition under which the mobile
network is unable to deliver normal service performance. Such events may result
from hardware or software faults, transmission failures, resource overload, or
planned and unplanned maintenance, and are associated with real service impact
experienced by subscribers~\cite{sterbenz2010resilience}.

A \emph{measurement artefact}, by contrast, refers to a spurious deviation caused
by errors in the collection, transmission, or aggregation of traffic counters.
Common sources include counter rollovers, partial or delayed reporting, and
transient monitoring faults. Although the underlying network may be operating
normally, such artefacts can produce apparent traffic drops that closely resemble
true failures, making them a major source of false alarms in traffic-based
monitoring systems.

Distinguishing between these phenomena is critical, as they produce similar signatures
in aggregated traffic data but require different operational responses. The proposed
framework is therefore designed to detect deviations from expected traffic dynamics
while facilitating separation of genuine failures from benign measurement artefacts.

Let $y_t \in \mathbb{R}_{\geq 0}$ denote the observed downlink traffic volume for a
given region during the $t$-th 15-minute interval, and let $x_t$ denote the
associated vector of temporal and contextual features. Observations arrive
sequentially and form a time-ordered data stream
$\{(x_t, y_t)\}_{t=1}^{\infty}$.

The objective is to learn a model that estimates the expected, or \emph{normal},
traffic dynamics $\hat{y}_t = f(x_t)$ and to identify observations whose
deviations from this expectation are inconsistent with normal operation. Such
deviations may arise from genuine network failures, benign measurement artefacts,
or transient contextual effects.

Unlike batch learning, the data generation process is inherently sequential:
measurements must be processed in arrival order, and future observations are
unavailable at training time. Models are therefore trained and updated
incrementally using online learning techniques, which more accurately reflect
operational deployment conditions than offline training followed by static
evaluation~\cite{ShalevShwartz2012Online,Gama2014Survey}.

An important practical consideration is data quality. Missing, invalid, or
corrupted measurements can bias learned baselines and degrade detection
performance. The proposed framework therefore supports selective learning, where
model updates are restricted to observations whose quality can be assessed with
confidence.

\subsection{Limitations of Threshold-Based Monitoring}
Threshold-based monitoring remains widely used in operational networks due to its
simplicity and low computational cost~\cite{Lakhina2004Characterization,Chandola2009Anomaly}.
Alarms are triggered when observed traffic exceeds predefined bounds, typically
configured using historical averages or expert knowledge.

Such approaches implicitly assume that acceptable traffic ranges are static or slowly varying, an assumption routinely violated in practice. Mobile traffic exhibits strong diurnal and weekly cycles, calendar effects, and long-term growth, causing fixed thresholds to generate false alarms during predictable low-usage periods while missing genuine anomalies during peak demand. Moreover, maintaining threshold-based systems requires continuous manual tuning,
which does not scale well across regions, technologies, and aggregation levels.

\subsection{Learning-Based Baseline Modelling}

The approach proposed in this work replaces static thresholds with a learned,
time-adaptive baseline. Rather than evaluating traffic against fixed limits,
anomalies are assessed relative to model expectations $\hat{y}_t$ that evolve over
time, allowing the system to accommodate non-stationarity, long-term trends, and
recurring temporal patterns without manual reconfiguration.

Anomaly detection is formulated as a streaming prediction problem in which
deviations between $y_t$ and $\hat{y}_t$ define an anomaly score. Continuous
incremental learning enables automatic adaptation to changing traffic conditions
and reduces reliance on heuristic rules, consistent with modern approaches to
non-stationary time-series analysis~\cite{Laptev2015Generic,Ren2019TimeSeries}.

A central challenge is extreme class imbalance: normal operation and measurement
artefacts occur frequently, while confirmed network failures are rare and labels
are often delayed. The proposed two-stage architecture explicitly addresses this
by decoupling self-supervised baseline traffic modelling from failure
classification. Normal dynamics is learned from abundant streaming data, while
failure detection operates on residuals and auxiliary indicators in a
cost-sensitive, semi-supervised manner. This design mitigates class imbalance,
reduces dependence on scarce labels, and aligns the learning process with
operational network workflows.

\section{Dataset Description}
\label{sec:dataset}
\subsection{Overview}

\begin{figure}[!t]
    \centering
    \includegraphics[width=0.95\linewidth]{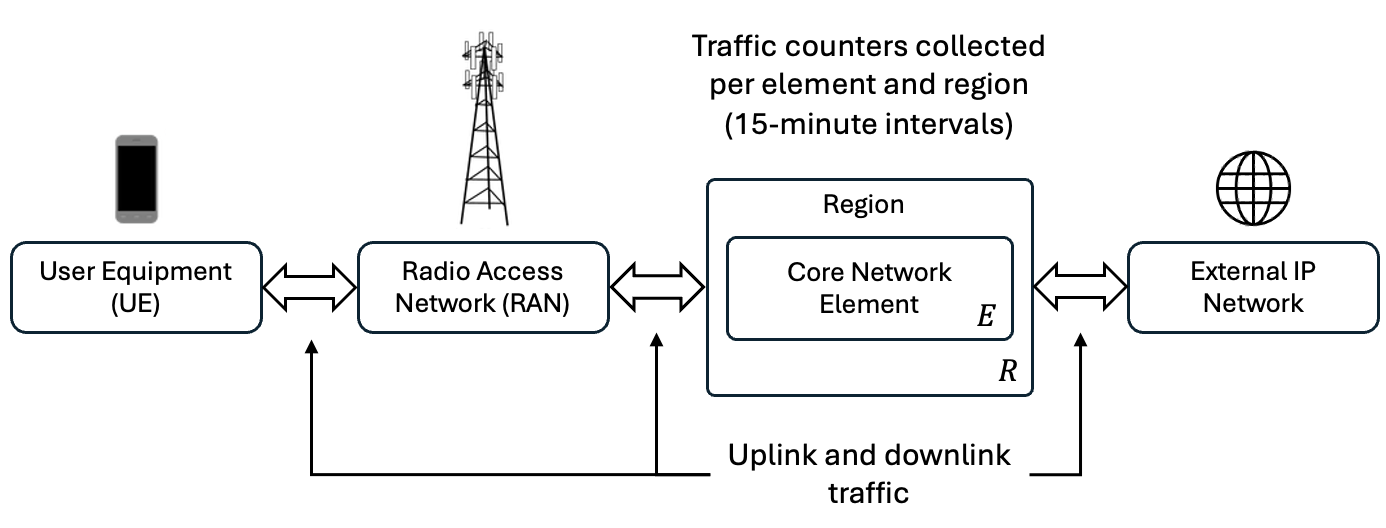}
    \caption{Aggregated traffic flow in an MCN. Traffic counters are
    collected per core network element $E$ and aggregated per geographical region $R$ at
    fixed 15-minute intervals.}
    \label{fig:core_traffic_flow}
\end{figure}

Figure~\ref{fig:core_traffic_flow} illustrates a basic flow of user traffic through a
mobile cellular network and the point at which measurements are collected for
monitoring. User traffic originates at the user equipment (UE), traverses the
radio access network (RAN), and is routed through multiple core network elements
before reaching external internet protocol (IP) networks. Both uplink and downlink traffic follow
this path in opposite directions.

For operational monitoring, traffic counters are collected at individual core
network elements and aggregated per geographical region at fixed 15-minute
intervals. The total observed traffic volume for a region therefore represents
the aggregate traffic routed through all reporting elements in that region during
the interval.

Traffic may traverse multiple core network elements within a region depending on
routing, and the set of reporting elements can vary over time due to equipment
failures, reconfigurations, or measurement artefacts. As a result, aggregated
traffic reflects both underlying user demand and the availability of the reporting
infrastructure. This coupling motivates a learning-based monitoring approach that
explicitly accounts for infrastructure dynamics when interpreting traffic
deviations.

The data\footnote{To protect commercially sensitive information, the dataset was anonymised and rescaled in a manner that preserves relative trends and temporal behaviour but prevents recovery of absolute traffic values.} used in this study was collected during routine network performance monitoring and analysed as part of a continuous improvement initiative within Vodacom Group Limited’s core network performance management function. The dataset consists of aggregated downlink traffic volume measurements with each observation corresponding to a region and time interval.

Let $y_{r,t}$ denote the observed downlink traffic volume for region $r$ at time
$t$. Traffic is obtained by aggregating counters across all network elements
$e \in E$ that successfully report during the interval:
\begin{equation}
    y_{r,t} = \sum_{e=1}^{E_{r,t}} y_{e,r,t},
\end{equation}
where $E_{r,t}$ denotes the number of reporting elements in region $r$ at time $t$.
Variations in $E_{r,t}$ due to incomplete reporting or monitoring faults can induce apparent traffic drops even when the network is operating normally.

The dataset is structured as a time-ordered data stream, consistent with the
operational setting in which measurements are generated continuously and processed
in near real time, making it well suited for evaluating online learning methods.

Ground-truth network failure labels were obtained from operational incident and fault
management records maintained by the MCN team. A failure label was
assigned to a region and time interval when a confirmed incident affecting service
delivery was recorded and temporally aligned with the corresponding traffic
measurement window. Confirmation typically followed manual investigation and root
cause analysis by network engineers.

As a result, labels are subject to practical limitations, including delay, partial
observability, and minor temporal misalignment between incident logs and traffic
aggregation intervals. The proposed framework explicitly accounts for this
uncertainty by decoupling self-supervised baseline learning from failure
classification and supporting asynchronous incorporation of confirmed labels.

\subsection{Features}

For each 15-minute interval, the following features are available:
\begin{itemize}
    \item \textbf{Downlink traffic volume} (\texttt{bytesDL}): total downlink traffic
    observed during the interval; the primary target variable.
    \item \textbf{Number of network elements} (\texttt{noElements}): number of core
    elements contributing counters during the interval.
    \item \textbf{Number of access point names} (\texttt{noAPNs}): number of APNs from
    which traffic was successfully recorded.
    \item \textbf{Geographical region} (\texttt{region}): categorical region identifier.
    \item \textbf{Timestamp} (\texttt{dateTime}): start time of the 15-minute interval.
\end{itemize}

\subsection{Infrastructure Effects on Traffic Measurements}

Core network elements include routers, gateways, and virtualised network functions
that collectively enable packet transport and service delivery~\cite{3gpp_core_network}.

At the time of data collection, two classes of elements were present. Legacy
elements were implemented as monolithic appliances and counted as a single
reporting unit, while modern elements were virtualised using network function
virtualisation (NFV), where a single logical function could comprise multiple
virtual machines, each contributing a separate counter.

As a result, missing statistics from a legacy element reduce the reported element
count by one, whereas missing statistics from a virtualised function may reduce
the count by several tens simultaneously. In both cases, the loss of counters
induces an apparent traffic drop even if the network remains operational.
The relationship between aggregated traffic volume and \texttt{noElements} is inherently non-linear, as legacy elements contribute traffic in approximately proportional units, whereas virtualised functions may remove multiple counters simultaneously, producing disproportionate traffic deviations that do not reflect true service degradation.

\subsection{Access Point Names}

An access point name (APN) defines how user traffic is routed from the mobile core
to external networks. Public APNs provide Internet access, while corporate APNs
connect users to private enterprise networks.

In the dataset, most elements served a small number of public APNs, which carried the majority of traffic volume. A subset of elements supported
hundreds of corporate APNs, which individually contributed relatively little
traffic. Consequently, missing statistics from public-APN-heavy elements produce
large apparent traffic drops with minimal change in \texttt{noAPNs}, whereas
missing corporate APN statistics lead to large drops in \texttt{noAPNs} with only
minor traffic reduction. This joint relationship is non-linear, meaning similar changes in reporting counts can have very different traffic effects.

\subsection{Implications for Anomaly Detection}

Aggregated traffic drops may arise from either genuine service-impacting failures
or inconsistencies in the measurement process. Because both produce similar
signatures in traffic volume alone, reliable anomaly detection requires models
that jointly consider traffic levels, infrastructure availability, and contextual
features.

This motivates the learning-based approach adopted in this work, where traffic
predictions are conditioned not only on temporal patterns but also on auxiliary
indicators such as \texttt{noElements} and \texttt{noAPNs}. Incorporating this
context enables more reliable separation of true network failures from benign
measurement artefacts.

\section{Feature Engineering}
\label{sec:features}
Aggregated mobile core traffic exhibits strong periodic structure and non-stationarity driven by human activity, calendar effects, and long-term growth. Capturing these dynamics in an online setting requires feature representations that preserve temporal continuity while remaining computationally efficient and
numerically stable. We therefore evaluate a set of time-aware feature representations
that explicitly encode periodicity and are compatible with incremental learning,
analysing the trade-offs between expressivity, dimensionality, and generalisation under
streaming constraints.
Because the dataset contains irregular missing observations, explicit lagged features are omitted.

\subsection{Temporal and Contextual Traffic Patterns}
\label{subsec:temporal_patterns}

Operational traffic exhibits repeatable diurnal, weekly, and monthly patterns.
Usage typically increases at the start of billing cycles, peaks during evening
hours, and reaches minima in the early morning. Tariff structures and calendar
effects, such as weekends, public holidays, and school holidays, introduce
systematic but benign variations that differ across regions.

Although predictable, these patterns can resemble network faults when interpreted
using static thresholds. We therefore incorporate explicit time-aware features,
including cyclic encodings of hour-of-day and day-of-week, calendar indicators,
and trend variables, enabling models to distinguish contextual variation from
abnormal behaviour.

\subsection{Categorical Time Encoding}

A simple approach treats temporal variables as categorical and applies one-hot
encoding. While this allows linear models to learn arbitrary non-monotonic
relationships and yields strong predictive performance, feature dimensionality
scales poorly with temporal resolution and ignores the cyclic structure of time,
increasing overfitting risk in streaming settings.

\subsection{Trigonometric Encoding}
Trigonometric encoding represents periodic time variables using sine and cosine functions, ensuring that the cyclic nature of time is preserved. This compact and numerically stable representation is well suited to online learning, but its global sinusoidal structure can underfit asymmetric or sharply varying traffic patterns.

\subsection{Periodic Spline Encoding}

Spline-based encodings provide increased expressivity while enforcing smoothness~\cite{eilers1996flexible}.
A periodic spline basis with $K$ components allows models to focus on specific
regions of the cycle while maintaining continuity. This approach offers
performance comparable to one-hot encodings with fewer features and improved
generalisation.

\subsection{Cyclic Radial Basis Function Encoding}

To encode periodic temporal variables in a smooth and locally adaptive manner,
we employ cyclic Gaussian radial basis function (RBF) encodings. Let
$x_t \in \{0,\ldots,P-1\}$ denote the discrete position of a time index within a
period of length $P$ (e.g., the 15-minute interval index within a day, where
$P=96$). A set of $K$ RBF centres $\{c_k\}_{k=1}^{K}$ is placed uniformly over the
periodic domain.

Cyclic distance between a time index $x_t$ and a centre $c_k$ is defined as:
\begin{equation}
    d_{\mathrm{cyc}}(x_t, c_k) =
    \min\bigl(|x_t - c_k|,\; P - |x_t - c_k|\bigr),
\end{equation}
which respects the wrap-around structure of periodic time.

Each basis function is computed using a Gaussian kernel:
\begin{equation}
    \phi_k(x_t) = \exp\!\left(-\frac{d_{\mathrm{cyc}}(x_t, c_k)^2}{2\sigma^2}\right),
\end{equation}
where $\sigma$ controls the width and smoothness of the encoding. The resulting
feature vector $\boldsymbol{\phi}(x_t) \in \mathbb{R}^K$ provides a bounded,
continuous representation of time that is well suited to incremental learning.

\subsection{Calendar and Trend Features}

To capture long-term drift and calendar effects, we include an ordinal date
feature, a first-day-of-month indicator, and a public holiday flag. Unbounded
features are standardised using running statistics, while bounded and binary
features are left unscaled.



\section{Two-Stage Online Learning Architecture}
\label{sec:online}

Aggregated core network traffic is treated as a \emph{reactive data stream}, where
observations arrive sequentially and must be processed in real time. At each time
step, predictions are generated using information available up to time $t-1$,
followed by incremental model updates once the true observation becomes available.
This \emph{predict--then--update} protocol mirrors operational deployment and avoids batch retraining.

The proposed framework follows a two-stage online learning architecture. Stage~I
incrementally learns a baseline model of normal traffic dynamics, while Stage~II
analyses deviations from this baseline, together with contextual indicators, to detect network failures while suppressing false alarms caused by benign variation and measurement artefacts.

\subsection{Stage I: Online Traffic Nowcasting}

Let $y_t$ denote the observed traffic volume and $\mathbf{x}_t$ the associated feature
vector at time $t$. Stage~I predicts the expected traffic
$\hat{y}_t = f_t(\mathbf{x}_t)$ and computes the residual
$e_t = y_t - \hat{y}_t$. Model updates occur strictly after prediction, following a
prequential (test--then--train) protocol. Models are selected for low computational
cost, stability, and continuous online operation.

\subsection{Stage II: Online Anomaly Assessment}

Stage~II evaluates whether deviations from the learned baseline correspond to genuine
network failures or benign measurement artefacts. At each time step, Stage~II receives
the signed residual $e_t$ together with a contextual feature vector
$\mathbf{z}_t$ comprising auxiliary indicators such as \texttt{noElements} and
\texttt{noAPNs}. A binary classifier $g_t(\cdot)$ produces a failure decision
$\hat{c}_t = g_t(e_t, \mathbf{z}_t)$, where $\hat{c}_t \in \{0,1\}$ denotes normal
operation or network failure, respectively.

Rather than applying fixed thresholds, classification is performed relative to recent
error statistics, improving robustness to gradual traffic evolution and extreme class
imbalance. The use of signed residuals preserves directional information, with negative values being more indicative of network failures. 


\subsection{Delayed and Sparse Labels}

In operational settings, network failure labels are often delayed or unavailable due
to manual investigation and root-cause analysis workflows. The proposed architecture
accommodates this constraint by decoupling the learning stages. Stage~I operates
fully self-supervised after an initial bootstrapping phase, during which a small
set of traffic observations $y_t^{(0)}$ corresponding to known measurement artefacts
is identified and excluded from model updates.

Stage~II operates in a semi-supervised manner and incorporates
confirmed failure labels asynchronously when they become available. A limited set
of initial network failure labels $c_t^{(0)}$ is used during bootstrapping, after which
additional confirmed labels $c_t^{+}$ are integrated over time as investigations
conclude. 

\refstepcounter{algorithm}
Algorithm~\ref{alg:two_stage} summarises the proposed framework, highlighting the
decoupling between self-supervised baseline learning and delayed, cost-sensitive
failure classification under a prequential evaluation protocol.

\vspace{0.5em}
\hrule
\vspace{0.3em}
\noindent\textbf{Algorithm \thealgorithm}
\label{alg:two_stage}
\vspace{0.3em}
\hrule
\vspace{0.5em}
\begin{algorithmic}[1]
\REQUIRE Streaming observations $\{(\mathbf{x}_t, y_t, \mathbf{z}_t)\}_{t=1}^{\infty}$
\REQUIRE Initial baseline model $f_0$, failure classifier $g_0$
\REQUIRE Warm-up period $T_0$
\STATE Initialise Stage~I model $f \leftarrow f_0$
\STATE Initialise Stage~II model $g \leftarrow g_0$
\STATE Initialise running statistics
\FOR{$t = 1,2,\ldots$}
    \STATE Receive features $\mathbf{x}_t$ and contextual indicators $\mathbf{z}_t$
    \STATE Predict baseline traffic $\hat{y}_t \leftarrow f(\mathbf{x}_t)$
    \STATE Observe true traffic $y_t$
    \STATE Compute residual $e_t \leftarrow y_t - \hat{y}_t$
    \STATE Predict failure decision $\hat{c}_t \leftarrow g(e_t, \mathbf{z}_t)$
    \IF{$t > T_0$ \AND observation quality acceptable}
        \STATE Update baseline model $f \leftarrow f(\mathbf{x}_t, y_t)$
    \ENDIF
    \IF{confirmed failure label $c_t^{+}$ becomes available}
        \STATE Update classifier $g \leftarrow g(e_t, \mathbf{z}_t, c_t^{+})$
    \ENDIF
\ENDFOR
\end{algorithmic}
\vspace{0.5em}
\hrule
\vspace{0.75em}

\subsection{Online Learning Implementation}

All models are implemented using the \emph{River} library~\cite{montiel2021river},
which supports incremental preprocessing, model updates, and prequential evaluation.
Deep neural network models are implemented using \emph{deep-river}~\cite{Kulbach2025},
a Python library for online deep learning that integrates the River API with
PyTorch-based neural network architectures.

Unbounded continuous features are standardised using running statistics maintained
online, while bounded or binary features are left unscaled. Model parameters are updated using stochastic gradient descent or Adam-based
incremental optimisers.

\subsection{Operational Considerations}

\begin{figure}[!h]
    \centering
    \includegraphics[width=0.95\linewidth]{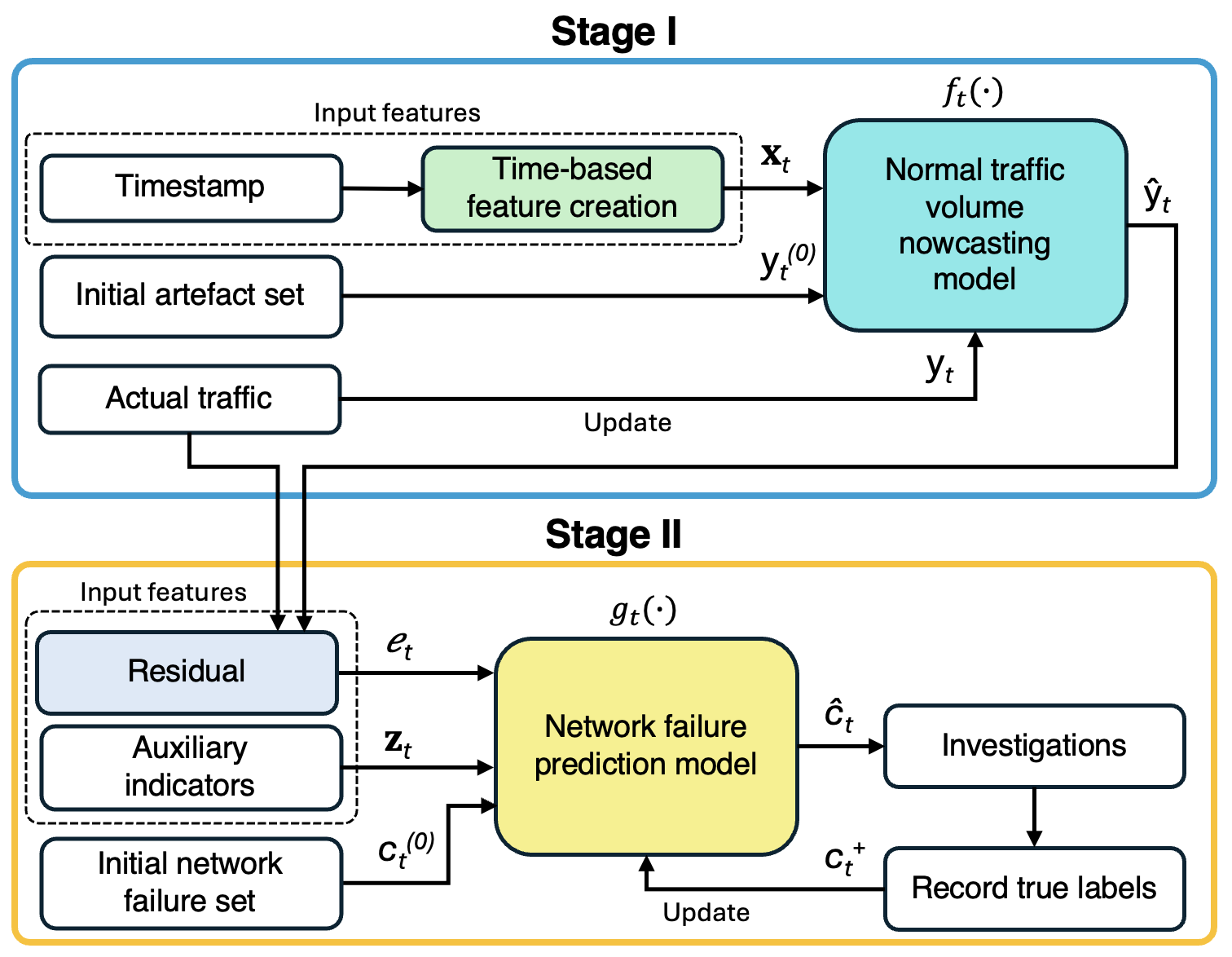}
    \caption{Two-stage online learning architecture for MCN traffic
monitoring. At each time step $t$, Stage~I predicts $\hat{y}_t$ from features
$\mathbf{x}_t$ and computes the residual $e_t$. Stage~II evaluates
residuals and contextual indicators to produce a failure decision $\hat{c}_t$, incorporating
delayed feedback ${c}^{+}_t$ when available.}
    \label{fig:architecture}
\end{figure}

Unlike batch-trained models that require periodic retraining, the proposed online
architecture adapts continuously to gradual traffic evolution, seasonal effects,
and concept drift. This results in reduced operational overhead, low computational cost, and minimal manual intervention, making the approach well suited to
large-scale mobile network monitoring. An overview of the two-stage architecture is shown in Fig.~\ref{fig:architecture}.

From a computational perspective, all components of the proposed framework are
lightweight. Stage~I employs an online linear regressor with 29 input features,
while Stage~II uses logistic regression with only three inputs. Both models
operate with constant memory and linear per-sample update complexity in the
number of features. In practice, Stage~I maintains fewer than 200 parameters per
region, while Stage~II introduces fewer than 10 additional parameters, making the
overall approach suitable for large-scale deployment across many regions with
minimal computational overhead.

\section{Experimental Results}
\label{sec:results}

This section evaluates the proposed two-stage online learning framework under
realistic streaming conditions. Experiments focus on three aspects:
(i) online traffic nowcasting accuracy, (ii) the impact of time-aware feature
representations, and (iii) failure detection performance for linear and non-linear models under severe class imbalance and measurement artefacts.

All models are evaluated using a prequential (test--then--train) protocol.
Unless stated otherwise, reported metrics are cumulative prequential values
computed after a 30-day warm-up period.


\subsection{Stage I: Online Traffic Nowcasting}
\label{subsec:stage1}

We first assess the ability of online regression models to learn and maintain a
baseline of normal traffic dynamics. Accurate baseline modelling is a prerequisite
for reliable anomaly detection and directly affects downstream failure detection
performance.

Table~\ref{tab:features} compares cumulative prequential prediction accuracy for
different temporal encodings using a fixed online linear regression model. One-hot
encodings of the 15-minute interval serve as a high-capacity baseline, while
trigonometric, periodic spline, and cyclic RBF encodings provide increasingly compact
and smooth representations of temporal structure.

\begin{table}[!h]
\centering
\caption{Impact of time-aware feature representations on online prediction accuracy.}
\label{tab:features}
\begin{tabular}{lcccc}
\toprule
\textbf{Encoding} 
& \makecell{\textbf{15-min}\\\textbf{Features}} 
& \makecell{\textbf{Total}\\\textbf{Features}} 
& \textbf{MAPE (\%)} 
& $\mathbf{R^2}$ \\
\midrule
One-hot & 96 & 142 & 6.15 & 0.97 \\
Trigonometric & 2 & 11 & 13.37 & 0.89 \\
Periodic spline & 16 ($K{=}16$) & 31 & 6.21 & 0.97 \\
Cyclic RBF & 12 ($K{=}12$) & 29 & \textbf{4.81} & \textbf{0.98} \\
\bottomrule
\end{tabular}
\end{table}

Cyclic RBF encodings achieve the best overall performance, delivering improved accuracy
with substantially fewer features. Periodic spline encodings capture asymmetric
diurnal patterns with performance comparable to one-hot representations, while
trigonometric features underfit the data due to their global sinusoidal constraint.
For completeness, the total number of model features associated with each encoding is
also reported, highlighting the efficiency gains of the cyclic representations.

Selective learning in Stage~I was implemented using a two-phase approach. Initially,
a small set of known measurement artefacts was manually identified and excluded from
model updates to establish a clean baseline.

While accurate baseline modelling is necessary, it is not sufficient for operational
network monitoring. In the following experiment, we evaluate how prediction residuals
produced by Stage~I can be leveraged to distinguish genuine network failures from
benign measurement artefacts under extreme class imbalance.

\subsection{Stage II: Failure Detection Performance}
\label{subsec:stage2}

Failure detection is evaluated under realistic operating conditions in which
confirmed network failures are rare and measurement artefacts occur frequently.
Performance\footnote{Due to the extreme class imbalance inherent in MCN monitoring data,
macro-averaged precision, recall, and F1-score are used as the primary evaluation
metrics, as micro- and weighted averages are dominated by normal operation and
can obscure failure detection performance.} is assessed using precision, recall, F1-score, false positive rate
(FPR), and area under the curve (AUC).

\begin{table}[!t]
\centering
\caption{Failure detection performance under extreme class imbalance.
All metrics are final cumulative prequential values obtained under a
test--then--train evaluation protocol.}
\label{tab:stage2}
\resizebox{\columnwidth}{!}{%
\begin{tabular}{lccccc}
\toprule
\textbf{Method} 
& \textbf{Precision} 
& \textbf{Recall} 
& \textbf{F1} 
& \textbf{FPR} 
& \textbf{AUC} \\
\midrule
\multicolumn{6}{l}{\textit{Threshold-based baselines}} \\
\midrule
Static threshold
& 49.3\% 
& 40.5\% 
& 44.2\% 
& 20.12\% 
& -- \\
Residual threshold
& 51.3\% 
& 50.8\% 
& 50.9\% 
& 0.64\% 
& -- \\
\midrule
\multicolumn{6}{l}{\textit{Linear models}} \\
\midrule
Single-model (linear)
& \textbf{92.6\%} 
& 61.5\% 
& 68.0\% 
& \textbf{0.04\%}
& 0.77 \\
Two-stage (linear)
& 74.0\% 
& 82.5\%
& \textbf{78.0\%} 
& 0.38\%
& 0.86 \\
\midrule
\multicolumn{6}{l}{\textit{Non-linear models}} \\
\midrule
Single-model (NN)
& 55.2\%
& 90.9\%
& 57.2\%
& 8.28\%
& 0.94 \\
Two-stage (NN)
& 63.7\%
& \textbf{96.6\%}
& 70.6\%
& 2.71\%
& \textbf{0.98} \\
\bottomrule
\end{tabular}%
}
\end{table}

Table~\ref{tab:stage2} compares static thresholding, residual-based thresholding, single-model learning, and the proposed two-stage online architecture. The term single-model denotes classifiers trained directly on traffic and contextual features, without explicit baseline modelling or residual extraction.

Static thresholding proves unsuitable under non-stationary traffic due to low precision, low recall, and high false positive rates, whereas residual-based thresholding reduces false alarms but provides only limited improvements in recall and overall F1-score.

Single-model learning achieves very high precision when linear classifiers are employed, but this conservative behaviour leads to reduced recall under extreme class imbalance. Introducing non-linear capacity via a feed-forward neural network increases recall markedly; however, this comes at the cost of a sharply elevated false positive rate, demonstrating that model expressiveness alone does not resolve the ambiguity between genuine failures and measurement artefacts.

In contrast, the two-stage architecture consistently delivers a more favourable precision–recall trade-off for both linear and non-linear models. By isolating predictable traffic dynamics in Stage~I, the residual-based representation simplifies the learning problem in Stage~II, improving recall while constraining false alarms. The linear two-stage model achieves the highest F1-score, while the two-stage neural network attains the highest AUC, confirming that explicit baseline decomposition is critical for reliable failure detection under streaming class imbalance. The modest increase in false positive rate for the two-stage linear model reflects its higher sensitivity to partial or shorter-duration failure events that the conservative single-model detector suppresses, a trade-off that yields substantially higher recall and overall F1-score.

\begin{figure}[!h]
\centering
\includegraphics[width=1.0\linewidth]{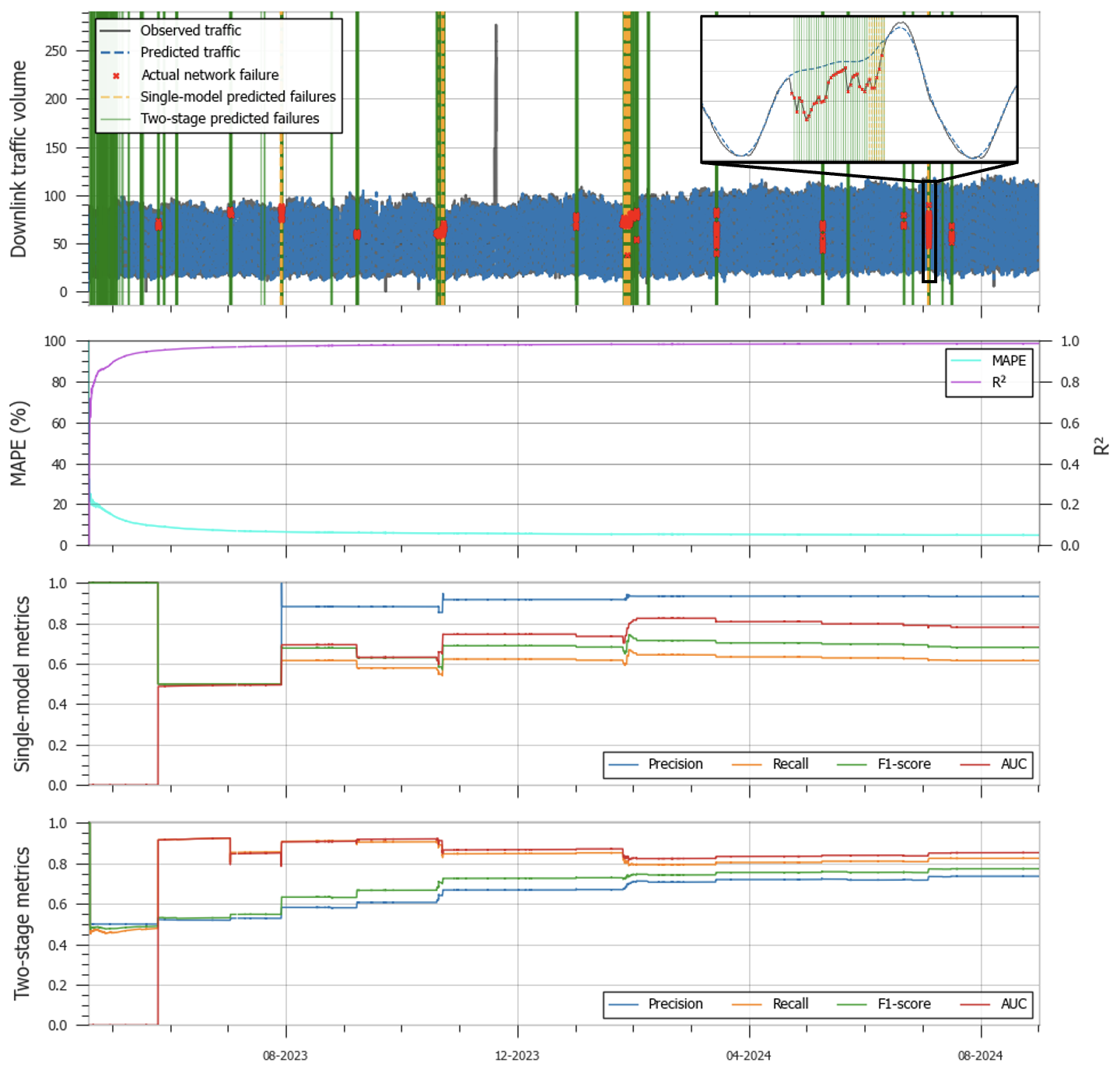}
\caption{Observed and predicted downlink traffic with
confirmed and predicted failure events. The upper panel shows
traffic dynamics and detection timing, while the lower panels show the prequential
evolution of model metrics. The two-stage model exhibits faster stabilisation and improved alignment
with true failure events while suppressing spurious detections.}
\label{fig:online_results}
\end{figure}

Figure~\ref{fig:online_results} provides a qualitative illustration of online failure detection behaviour for the linear models, complementing the aggregate results in Table~\ref{tab:stage2}. The upper panel shows observed and predicted downlink traffic with confirmed failure events and predicted detections overlaid, while the lower panels report the cumulative prequential evolution of precision, recall, F1-score, and AUC.

The single-model linear detector exhibits strongly conservative behaviour. Predicted failures occur infrequently and are tightly concentrated around the most severe traffic drops, resulting in high precision and a low false positive rate. However, several confirmed failure intervals are missed or only partially detected, causing recall to plateau at a lower level and constraining the achievable F1-score, as reflected in the metric trajectories.

In contrast, the two-stage linear model shows improved temporal alignment with confirmed failure events. By operating on residuals produced by the Stage~I traffic baseline, the detector responds more consistently throughout true failure periods while maintaining a similarly low false positive rate. This leads to a sustained increase in recall with only a modest reduction in precision, yielding a higher and more stable F1-score and improved AUC over time.

The zoomed inset highlights a representative failure episode, where the two-stage model generates contiguous detections aligned with the confirmed incident, while the single-model detector triggers sporadically or not at all. This visual evidence reinforces the quantitative findings and demonstrates that isolating predictable traffic dynamics in Stage~I simplifies downstream classification, improving failure sensitivity without compromising operational robustness.

\section{Conclusions and Future Work}
\label{sec:conclusion}

This paper presented a two-stage online learning framework for traffic-based failure
detection in MCNs. By explicitly separating baseline traffic modelling
from failure classification, the approach achieves reliable performance under
non-stationarity, extreme class imbalance, and imperfect telemetry.

Experiments on real operational MCN data demonstrate that the two-stage architecture
outperforms static thresholds and single-model detectors, delivering higher
F1-scores and AUC at low false positive rates.

The framework was developed and evaluated in collaboration with Vodacom Group Limited,
demonstrating its suitability for operational deployment. Future work will explore
adaptive model selection and extensions to multi-resolution and cross-region traffic
dependencies.

\bibliographystyle{IEEEtran}
\bibliography{references}

\end{document}